\documentclass{article}

\usepackage{algorithm}
\usepackage[noend]{algpseudocode}
\usepackage{amsmath}
\newcommand\ddfrac[2]{\frac{\displaystyle #1}{\displaystyle #2}}

\DeclareMathOperator*{\argmin}{\arg\!\min}
\usepackage[final]{pdfpages}
\usepackage{wrapfig}
\usepackage{graphicx}
\graphicspath{ {./images/} }

\usepackage[numbers, sort]{natbib}

\usepackage[final]{neurips_2022}

\usepackage[utf8]{inputenc} 
\usepackage[T1]{fontenc}    
\usepackage[pagebackref=true,breaklinks=true,colorlinks,urlcolor=blue,citecolor=green,linkcolor=green,bookmarks=false]{hyperref}
\usepackage{multirow}
\usepackage{subcaption}

\usepackage{url}            
\usepackage{booktabs}       
\usepackage{amsfonts}       
\usepackage{nicefrac}       
\usepackage{microtype}      
\usepackage{xcolor}         

\title{GenSDF: Two-Stage Learning of \\Generalizable Signed Distance Functions}

\author{
 Gene Chou \\
 Princeton University\\
 \texttt{gchou@princeton.edu} \\
   \And
   Ilya Chugunov \\
   Princeton University \\
   \texttt{chugunov@princeton.edu} \\
   \And
   Felix Heide \\
   Princeton University \\
   \texttt{fheide@princeton.edu} \\
}

\begin{document}

\maketitle

\begin{abstract}
We investigate the generalization capabilities of neural signed distance functions (SDFs) for learning 3D object representations for unseen and unlabeled point clouds. Existing methods can fit SDFs to a handful of object classes and boast fine detail or fast inference speeds, but do not generalize well to unseen shapes. We introduce a two-stage semi-supervised meta-learning approach that transfers shape priors from labeled to unlabeled data to reconstruct unseen object categories. The first stage uses an episodic training scheme to simulate training on unlabeled data and meta-learns initial shape priors. The second stage then introduces unlabeled data with \emph{disjoint classes} in a semi-supervised scheme to diversify these priors and achieve generalization. We assess our method on both synthetic data and real collected point clouds. Experimental results and analysis validate that our approach outperforms existing neural SDF methods and is capable of robust zero-shot inference on 100+ unseen classes. Code can be found \href{https://github.com/princeton-computational-imaging/gensdf}{here}.
\end{abstract}

\section{Introduction}
Learned 3D representations form the core of shape recovery~\cite{surface_recon_survey}, robotic manipulation~\cite{robotics_manipulation_planning}, scene understanding~\cite{semi-sup-scene}, and content generation~\cite{stylesdf} tasks. While simple to interpret, high-quality explicit 3D representations can come with a slew of hidden costs. Conventional voxel grids without acceleration suffer from cubical memory requirements~\cite{voxnet, octree, voxfields}, dense point clouds contain large amounts of redundant information make querying local structures computationally expensive~\cite{pointnet,pointnet++}, and direct optimization of mesh representations is often infeasible~\cite{atlasnet}. As an alternative, works such as~\cite{deepsdf, occupancy, nerf} propose \emph{implicit} learned representations; networks trained to functionally approximate some underlying 3D geometry. These methods can offer lower memory requirements and faster training and inference speeds~\cite{singleshape} than conventional explicit representation methods~\cite{voxnet,pointnet}.

Neural signed distance functions (SDFs)~\cite{deepsdf} are a subset of this work which implicitly model the distance from a queried location to the nearest point on a shape's surface -- negative inside the shape, positive outside, zero at the surface. Existing methods are capable of approximating diverse synthetic geometry, including multiple objects~\cite{deepsdf, metasdf, conv_occ} and varying levels of detail for single objects~\cite{bacon, nglod}. However, most successful methods are fully supervised--they require access to ground truth signed distance values to fit to an object. As ordinary point clouds lack these values, a fully-supervised approach cannot learn from such unlabeled data.

To replace the need for any annotated data, fully unsupervised approaches~\cite{neuralpull, sal, sald} operate directly on point clouds. These methods leverage pseudo-labels and proxies for ground truth signed distance values, for example the unsigned distance to the nearest point in the point cloud. Unfortunately, existing unsupervised methods can only reconstruct a handful of shapes, and break down when we introduce just three or four additional additional object categories. As a result, these approaches can also not take advantage of large unlabeled point cloud datasets~\cite{semantic3d, ycb, scannnobj} and their real-world applications are limited. Our proposed method lifts this limitation, and learns from large-scale unlabeled data, successfully reconstructing an order of magnitude more categories in inference than seen during training.

To achieve this we propose a semi-supervised learning approach. Existing semi-supervised methods use labeled and unlabeled data of the same class to improve intra-class performance for tasks such as 3D reconstruction from images~\cite{semi-sup-from-img, semi-sup-3dhumanpose, semi-sup-face} and scene reconstruction from point clouds~\cite{semi-sup-scene, siren}. Although they perform well on seen data, these methods are trained on labeled and unlabeled data of the same class, and so do not by themselves generalize well to unseen classes. We propose a novel semi-supervised meta-learning learning approach that learns geometric relations between labeled and unlabeled data of non-overlapping categories. One which learns an SDF that generalizes to a diverse set of unseen, out-of-distribution geometries.

We train in two stages. In the first meta-learning stage, we learn shape priors from the labeled set. Different from existing supervised methods, we introduce an episodic training scheme where we split data into subsets with disjoint classes to learn a generalized shape representation. One part of this split is treated as a labeled supervised set of data, with ground truth signed distance penalties, while the other is used to emulate unlabeled data samples, with only self-supervised losses. We validate that this first stage provides a generic shape prior which favors expressiveness at the cost of reconstruction quality. In the second, semi-supervised stage, we introduce a new fully disjoint set of unlabeled data to the training scheme. Successful training hinges on our novel self-supervised sign prediction loss, which diverges from the unsigned convention of previous works~\cite{neuralpull, sal, sald}. This second stage allows the model to ingest large amounts of unlabeled data to diversify the shape model learned in the first stage, which significantly increases the quality of the signed distance predictions over a wide set of unseen shape classes.

We evaluate our method on Acronym~\cite{acronym}, a synthetic dataset, and YCB~\cite{ycb}, a collection of real-world point clouds. We validate the generalization capabilities of our approach, and find that it outperforms state-of-the-art learned SDF methods and is capable of robust zero-shot inference~\cite{ZSL_survey} on 100+ unseen classes and scanned point clouds.

\section{Related Work}
\label{related_work}

\paragraph{Learning Implicit Surface Representations.}
Implicit neural representations form the backbone of many modern 3D reconstruction and view synthesis works~\cite{deepsdf, occupancy, nerf, hndr,nerfinthewild}. They use neural networks as \emph{universal approximators}~\cite{mlp} to directly learn a mapping between spatial or spatio-temporal~\cite{nerfies} coordinates and scene attributes, which offers a fully-differentiable and space-efficient~\cite{siren} way to represent 3D geometry. Park et al.~\cite{deepsdf} first experimented with mapping coordinates to signed distance values -- a neural signed distance function (SDF) -- whereby an object is implicitly represented by the distances between 3D coordinate queries their closest surface point.

With labeled data -- when all 3D query points are annotated with ground truth signed distance values -- predictions of signed distance values can be supervised directly~\cite{deepsdf, metasdf}. Existing methods~\cite{deepsdf,metasdf} require test-time fine-tuning even for a single class, prohibiting real-time inference or prediction on unlabeled data, or overfit to single objects~\cite{bacon, nglod}. Occupancy-based methods~\cite{conv_occ, if-net, localgrid} can learn a rudimentary shape prior, but still struggle to reconstruct fine features in unseen shapes. The proposed semi-supervised method learns a robust shape prior, and substantially outperforms fully supervised methods when tested on a set of over one hundred unseen categories.

With only unlabeled data -- un-annotated point clouds --  approaches try to generate proxies for signed distance values for supervision. Ma et al.~\cite{neuralpull} look at the unsigned distance to the nearest point in the point cloud as a proxy, and Atzmon and Lipman~\cite{sal} propose a class of loss functions and geometric initializations such that these unsigned distances converge to signed values. We propose a signed nearest neighbor loss and demonstrate that our approach, unlike existing unsupervised methods, is able to reconstruct thin object structures and converge on a dataset with large object variance.

\paragraph{Semi-supervised Surface Representations.} Traditional semi-supervised methods rely on learning from datasets where each category contains labeled and unlabeled samples. During inference they rely on pseudo-labels~\cite{semi-sup-label, semi-sup-label2}, or employ ensemble techniques to enforce consistency~\cite{semi-sup-retrieval}. These methods cannot be directly applied to target data from unseen categories. 

For 3D representation tasks, existing methods~\cite{semi-sup-from-img, semi-sup-3dhumanpose, semi-sup-face} use semi-supervision to reconstruct geometries from images or point clouds~\cite{semi-sup-scene, siren}. Elich et al.~\cite{semi-sup-from-img} pretrain their model with labeled data to learn 3D representations, to recover known shapes from RGBD data. Li et al.~\cite{semi-sup-scene} use labeled data as dense boundary values for semi-supervised SDF learning, and fit road scenes from point clouds. While these approaches improve predictions on seen data, they cannot generalize well to unseen classes, even with test-time optimization~\cite{semi-sup-scene}. Recently, person re-identification and domain adaptation methods perform semi-supervision on non-overlapping labeled and unlabeled classes and have shown to generalize better~\cite{semi-sup-cite, semi-gen1, semi-gen2}. This has not been attempted for learning SDFs, and our method introduces a meta-learning approach to learn shape priors, where we emulate training on unlabeled data with pseudo-labels. 

\paragraph{Generalization.} 
The ability to learn generalizable 3D representations is necessary for real-world applications, as real objects exhibit significantly more diversity than any dataset. In learning implicit functions, one approach~\cite{PatchNet, LDIF} is to break down shape features to find common properties, but this also requires prior knowledge about target shapes which might not be available. 

We propose a meta-learning approach that does not make assumptions or require prior information about target data. Meta-learning has been extensively studied in few-shot learning~\cite{few-shot1, few-shot2}, domain generalization~\cite{metareg}, and image retrieval~\cite{semi-sup-retrieval}, 3D reconstruction from 2D or 2.5D images~\cite{zeroshot_img, fewshot_img, fewshot_img2}, and point cloud classification and segmentation tasks~\cite{zeroshot_pc, fewshot_pc, fewshot_pc2}. However, meta-learning has not been extensively investigated in the context of SDFs. The most relevant method is MetaSDF~\cite{metasdf}, albeit it is fully supervised and exhibits large time and memory complexity due to inner loop gradient descent steps. Furthermore, MetaSDF overfits to single objects and requires ground truth signed distance values during test time. The proposed episodic training approach avoids this costly inner loop, and learns a unified network that only requires a raw input point cloud for reconstruction.

\section{Semi-supervised Meta-Learning of Implicit Functions}
\label{method}
We next formalize the problem we are solving in Sec.~\ref{sec:problem}, then describe the proposed learning method and training objectives in Sections~\ref{sec:stage1} and \ref{sec:stage2}.  

\begin{figure}[t]
    \centering 
    \includegraphics[width=\textwidth]{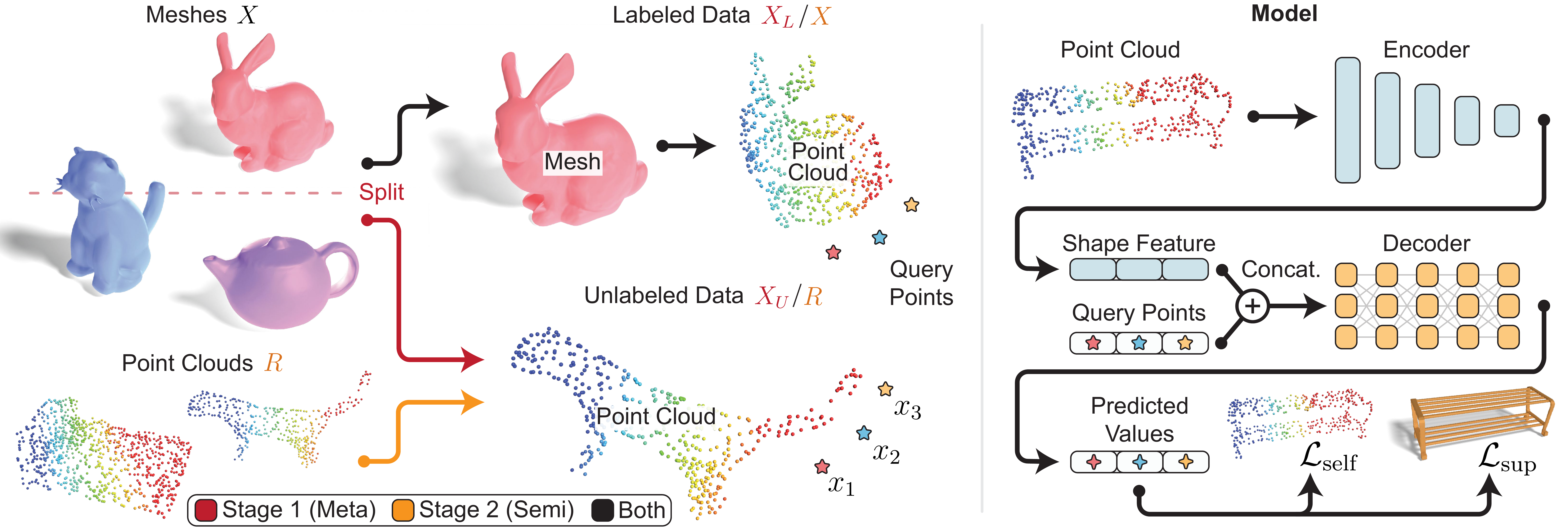}
    \caption{Our two-stage learning pipeline. The first stage (\textbf{red}) episodically splits the dataset into two subsets of disjoint categories of emulated labeled and unlabeled data, trained with combined supervised $\mathcal{L}_\mathrm{sup}$ and self-supervised $\mathcal{L}_\mathrm{self}$ loss. The second stage (\textbf{orange}) introduces real unlabeled data with non-overlapping classes in a semi-supervised fashion to improve model generalization.}
    \label{fig:training}
\end{figure}

\subsection{Problem Formulation}\label{sec:problem}
\paragraph{Conditional Signed Distance Functions}
We are interested in learning a generalized representation for shapes.
We learn a signed distance function (SDF), which is a continuous function that represents the surface of a shape approximated by the zero level-set of a neural network. Given a raw input point cloud $P = \{p_i \in \mathbb{R}^3 \}_{i=1}^N$ with $N$ points and a 3D query point $x \in \mathbb{R}^3$, we learn a conditional SDF $\Phi : \mathbb{R}^3 \times \{p_i \in \mathbb{R}^3 \}_{i=1}^N \rightarrow \mathbb{R}$ such that the function $\Phi$ predicts the signed distance value for a 3D coordinate conditioned on a sparse point cloud. Then $\Phi$ is a unified function that can be directly applied to any target point cloud. 
\vspace{-2.0mm}
\begin{equation}
  x, P \mapsto \Phi(x, P) = s,
  \label{eq:sdf-definition}  
\vspace{-1.5mm}
\end{equation}
where $s$ denotes the predicted signed distance value between $x$ and the shape described by $P$. 

The surface boundary of a shape described by $P$ is its zero-level set $S_0(\Phi(P))$, which can be expressed as
\begin{equation}
    S_0(\Phi(P)) = \{z\in \mathbb{R}^3\ |\ \Phi(z, P)=0 \}. 
    \label{eq:sdf-zero-level-set}
\end{equation}
Thus, given a trained $\Phi$, we can visualize the surface of some $P$ by drawing its zero-level set. 

\paragraph{Problem Setting}
We assume both labeled and unlabeled datasets are available, denoted $X$ and $R$, respectively. $X = \{P_{X}, \mathrm{SDF} \}$, where $P_{X}$ is a set of point clouds and $\mathrm{SDF}(\cdot)$ denotes the ground-truth SDF operator that is defined for all query points $x \in \mathbb{R}^3$. $R = \{P_R\}$, where $P_R$ is a set of point clouds belonging to classes disjoint with those of $P_X$. Each point cloud is a set of 3D coordinates that describes the surface boundary of an object, and contains no additional information such as normals.

To learn a generalizable conditional SDF, we propose a two-stage semi-supervised meta-learning algorithm.
In the first stage, we aim to learn shape priors using $X$.
As illustrated in Figure~\ref{fig:training}, we propose to train our model using an episodic scheme~\cite{metareg}, where every $f$ epochs we randomly split $X$ into two subsets with \emph{disjoint} categories.
In the second stage, as shown in Figure~\ref{fig:training}, we continue to train the pretrained weights from the first stage and train the model on $X$ and $R$ in a semi-supervised fashion.
We obtain a unified neural network $\Phi$ that learns from one dataset and can generalize to represent SDFs of objects in unseen categories during test time; i.e., zero-shot inference~\cite{ZSL_survey}.

\begin{wrapfigure}{rh}{0.5\textwidth}
 \vspace{-0.5em}
\centering
\includegraphics[width=0.5\textwidth]{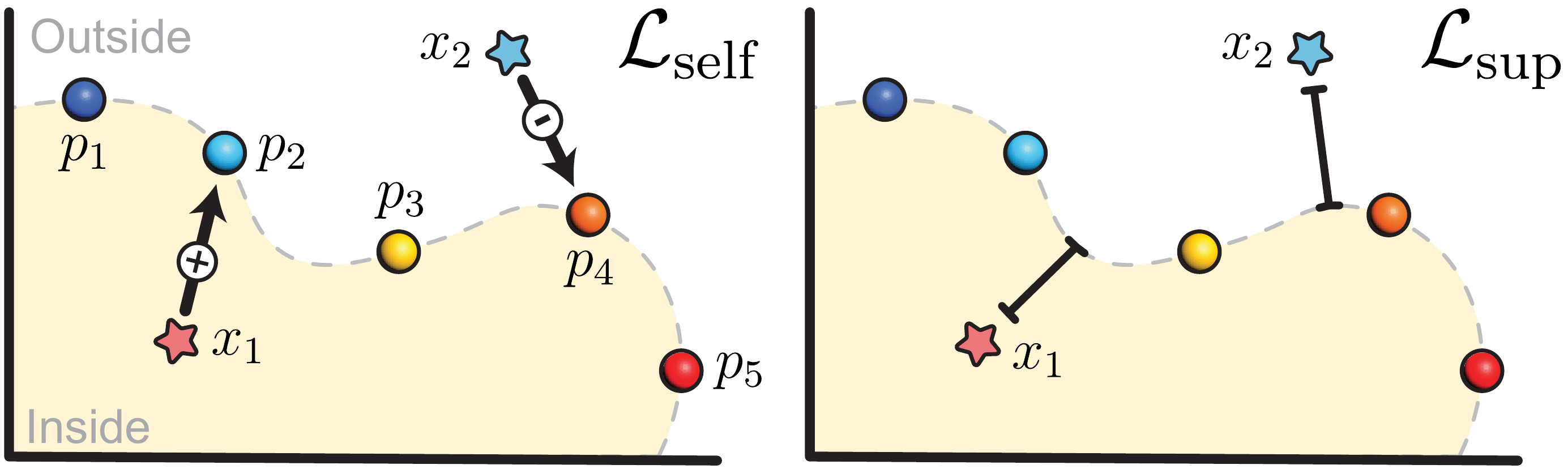}
 \caption{Our self-supervised (\textbf{left}) loss uses the nearest neighbor of a query point to approximate \emph{signed} distance. The supervised loss (\textbf{right}) instead penalizes based on exact signed distance.}
 \vspace{-0.5em}
\label{fig:losses}
\end{wrapfigure}

\subsection{Stage 1: Meta-Learning a Shape Prior}
\label{sec:stage1}
Our eventual goal is to train on labeled and unlabeled data simultaneously, so we mimic this goal in a meta-learning approach as we learn shape priors from $X$. Every $f$ epochs, we split $X$ into two subsets $X_L$ and $X_U$ with disjoint categories. We train $X_L$ with a supervised loss $ \mathcal{L}_{\mathrm{sup}}$, similar to typical supervised approaches on labeled data. We take away ground-truth signed distance values from $X_U$ making it ``pseudo-unlabeled'', and train it with our self-supervised loss $ \mathcal{L}_{\mathrm{self}}$. 
Our training objective is to minimize 
\begin{equation}
    \mathcal{L}_{\mathrm{meta}} = \mathcal{L}_{\mathrm{sup}}(X_L) + \lambda_{m} \mathcal{L}_{\mathrm{self}}(X_U)
    \label{metaeq}
\end{equation} where $\lambda_{m}$ controls the strength of the self supervised loss $ \mathcal{L}_{\mathrm{self}}$.
This meta-learning approach intends to accomplish two goals. First, the model learns to backpropagate with our self-supervised loss, which we need for training on unlabeled data in the second stage. Second, episodic training on non-overlapping categories forces the model to learn a generalized shape representation rather than overfit to geometries within a single category.

Compared to directly training on labeled and unlabeled data, this meta-learning process provides additional flexibility as we can tune the split frequency and ratio of labeled and ``pseudo-unlabeled'' data. The model repeatedly sees the same point clouds with and without labels, so shape features generated during labeled splits are shared with ``pseudo-unlabeled'' splits. In contrast, we find that semi-supervised training from scratch is unstable because errors in unlabeled data accumulate (see Tab.~\ref{table:ablation}). We empirically show in our ablation study that our method has stronger representation power for both seen and unseen classes. 
We provide an example of a training step in Alg.~\ref{metaalg} and a diagram in Fig.~\ref{fig:training}. Next, we explain our loss functions in more detail.

\paragraph{Supervised Loss Component}
In the supervised setting, we have the ground truth SDF operator; i.e., we know the signed distance values for all query points $x$. Therefore, the loss function is simply
\begin{equation}
    \mathcal{L}_{\mathrm{sup}} = \frac{1}{K}\sum_{k\in K} \| \Phi(x_k) - \mathrm{SDF}(x_k)\|_1 ,
    \label{supeq}
\end{equation}
\vspace{-6pt}
where $\| \cdot \|_1$ is the L1 loss.

\paragraph{Self-supervised Loss Component for Unlabeled Training Samples}

For each point $x \in \mathbb{R}^{3}$, we use its closest point $p \in P$ to approximate its projection to the surface, defined as
\begin{equation}
    t = \argmin_{p\in P} \|x-p\|_2 .
\end{equation}
In the literature~\cite{object_collision}, it is common to use 
\begin{equation}
    \hat{t} = x - \nabla \Phi (x) \Phi(x)
    \label{unsupeq}
\end{equation}
to approximate the closest point $t$ on the object surface to query point $x$ given some trained SDF function $\Phi$ for tasks such as collision detection. We build on this insight: we want to predict $\hat{t}$ to approximate $t$ on the point cloud by using the predicted signed distance value. However, we find that directly using Eq.~\eqref{unsupeq} leads to highly inaccurate sign predictions. There are no explicit penalties for predicting the wrong sign and when the query point is close to the surface, an $\ell_2$ loss between $\hat{t}$ and $t$ can be very low even when the sign is predicted incorrectly. For a surface boundary to be accurately estimated, there must be accurate sign changes between 3D coordinates.
\begin{algorithm}[t]
   \caption{Stage 1: Meta-Learning a Shape Prior}
    \begin{algorithmic}[1]
        \State \textbf{preprocess}: a labeled set $X$
        \State \textbf{initialize}: learning rate $\eta$, split frequency $f$, hyperparameter $\lambda_{m}$, model $\Phi$ with parameters $\theta$
        \State \textbf{for} $e$ \textbf{in range}(num\_epochs):
        \State \qquad \textbf{if} ($e$ \texttt{mod} $f$) == 0:
        \State \qquad \qquad  $X_L, X_U$ $\leftarrow$ \texttt{split}($X$)
        \vspace{0.2em}
        \State \qquad $\mathcal{L}_{\mathrm{sup}} = \frac{1}{|X_L|} \sum\limits_{(x,\text{SDF})\in X_L} \| \Phi(x;\theta^e), \text{SDF}(x) \|_1 $ \Comment{Eq. \eqref{supeq}}
        \vspace{0.2em}
        \State \qquad $\mathcal{L}_{\mathrm{self}} = \frac{1}{|X_U|} \sum\limits_{(x,t)\in X_U} \| (x \pm \frac{x-t}{\|x-t\|} \cdot \Phi(x;\theta^e)),\enspace t \|_2^2 $ \Comment{Eq. \eqref{selfeq}, simplified}
        \vspace{0.2em}
        \State \qquad $\mathcal{L}_{\mathrm{meta}} = \mathcal{L}_{\mathrm{sup}} + \lambda_{m} \mathcal{L}_{\mathrm{self}}$ \Comment{Eq. \eqref{metaeq}}
        \vspace{0.2em}
        \State \qquad $\theta^{e+1} \leftarrow \theta^e - \eta \nabla \mathcal{L}_{\mathrm{meta}} $
        \State \Return $\theta$
    \end{algorithmic}
    \label{metaalg}
\end{algorithm}

Thus, we substitute $\nabla \Phi (x)$ with the direction vector between $t$ and $x$. For incorrect sign predictions, the vector points in the direction opposite to the surface, guaranteeing a larger error. We denote our approach self-supervised because the predictions of the signs are used as labels. This allows us to leverage ground-truth signs during the meta-learning stage to guide training since the sign and absolute distance can be disentangled. In contrast, $\nabla \Phi (x) \Phi(x)$ is calculated as a unit that points in the direction of maximum distance change. Our training objective is to minimize the distance between $\hat{t}$ and $t$. Furthermore, we add an additional $\ell_1$ loss by using the points on point clouds as they have a distance value of $0$, which helps with thinner structures. Our final loss function is 
\begin{equation}
    \mathcal{L}_{\mathrm{self}} = \frac{1}{K} \sum_{k \in K} \| \hat{t}_k - t_k \|_2^2 + \lambda_{p} \frac{1}{N}\sum_{p\in P_{\mathrm{unlab}}} \| \Phi(p_n)\|_1,
    \label{selfeq}
\end{equation} 
where $\lambda_{p}$ determines the weight of the point cloud predictions and
\begin{equation}
    \hat{t} =\begin{cases}
x - \ddfrac{x-t}{\|x-t\|} \  \Phi(x)  & \text { if } \Phi(x) \ge 0, \\[2em]
x + \ddfrac{x-t}{\|x-t\|} \ \Phi(x)  & \text { if } \Phi(x) < 0 .
\end{cases}
\label{t_eq}
\end{equation} 
Previous work~\cite{neuralpull} has predicted $t$ by directly calculating Eq.~\eqref{unsupeq} during training, but their resulting sign predictions are substantially less accurate than ours for the reasons we discuss. We validate this with the results shown in Tab.~\ref{table:unlab}. We note that a point cloud  $P$ is sampled from a surface and does not represent the continuous boundary. Thus, $t$ is the nearest neighbor that exists in $P$ but not necessarily the true one that exists in the target surface we want to reconstruct. As a result there are inaccuracies and previous unsupervised work~\cite{neuralpull} that use $P$ as a proxy are not robust when training on multiple classes. Our two-stage approach lifts this constraint.

\begin{figure}[t]
    \centering
    \includegraphics[width=0.45\textwidth]{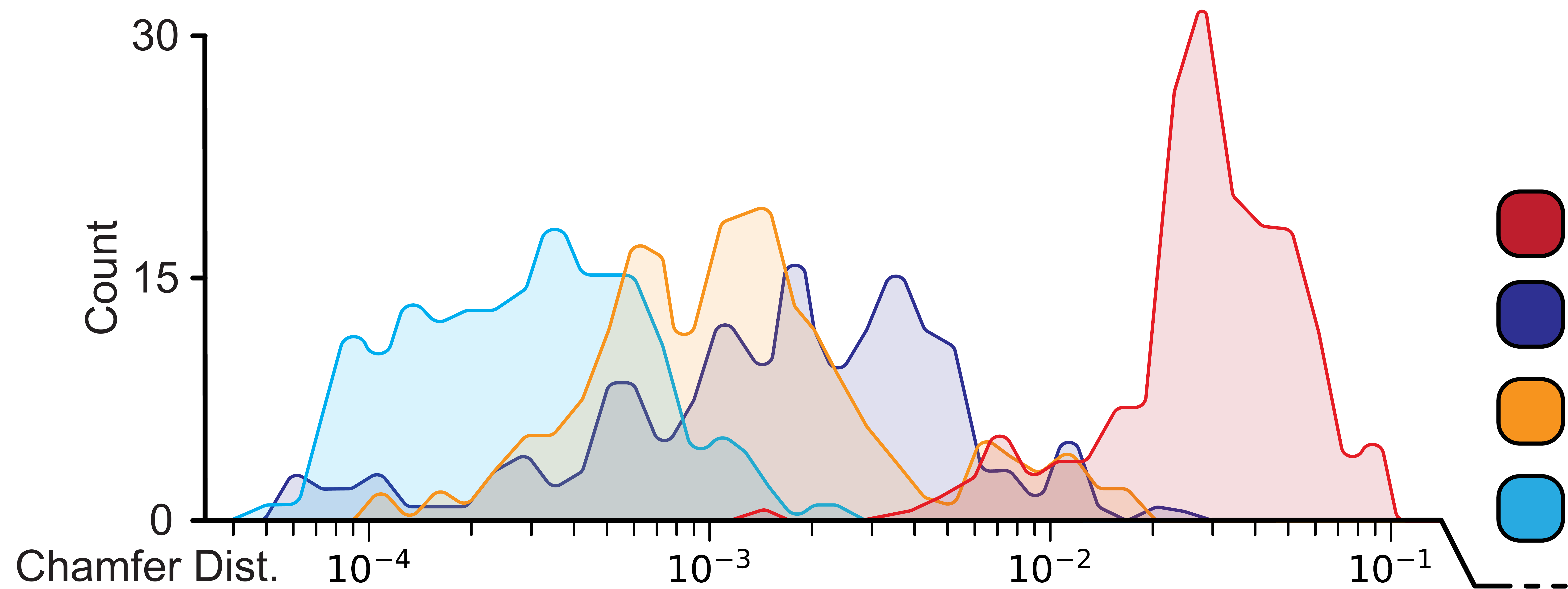}
    \begin{tabular}[b]{lcc}
    \toprule
          & Seen Classes & Unseen Classes\\
    \midrule
			\midrule
    DeepSDF & 18.26$/$10.87  & 33.19$/$29.74    \\
    MetaSDF  & 4.269$/$1.536  & 2.834$/$1.746      \\
    ConvOccNet  & 1.348$/$0.715  & 2.333$/$1.032      \\ 
    Proposed  & \textbf{0.351}$/$\textbf{0.166}  & \textbf{0.407}$/$\textbf{0.250}      \\ 
    \bottomrule
  \end{tabular}
    \caption{Mean/median Chamfer Distance (CD) of 20 seen and 166 unseen classes (\textbf{right}, scale is $10^{-4}$). For our method we report numbers prior to any test-time refinement. We include a histogram plot for visualizing the per-class CD distribution of the unseen classes (\textbf{left}, log scale). }
    \label{fig:hist}
    \vspace{-1em}
  \end{figure}

\subsection{Stage 2: Semi-supervision with Unlabeled Training Samples}\label{sec:stage2}

In our second stage, visualized in Fig.~\ref{fig:training}, we train our conditional SDF, pretrained in the meta-learning stage, on both labeled and unlabeled data in a semi-supervised fashion. As a result of the learned priors, training remains robust even with large amounts of unlabeled data. Furthermore, continuing to train with labeled data provides strong supervision to ensure inaccuracies from unlabeled data do not build up. We show in our ablation study (Sec.~\ref{sec:ablation}) that semi-supervised training from scratch can be unstable. Given labeled set $X$ and unlabeled set $R$, our training objective is to minimize
\begin{equation}
    \mathcal{L}_{\mathrm{semi}} = \mathcal{L}_{\mathrm{sup}}(X) + \lambda_{s} \mathcal{L}_{\mathrm{self}}(R)
    \label{semieq}
\end{equation} 
where $\lambda_{s}$ controls the strength of the self supervised loss $ \mathcal{L}_{\mathrm{self}}$. 

\section{Experiments}
\label{sec:experiments}

In this section, we analyze and validate the proposed representation method for seen and unseen object categories. We begin with a discussion of implementation details and experimental setup.

\paragraph{Implementation} For our experiments, we borrow the encoder architecture from~\cite{conv_occ} which is a modified PointNet~\cite{pointnet} that uses plane projection to learn local geometric features, and parallel UNets~\cite{UNet} to aggregate spatial information. For the decoder we use an 8-layer multi-layer perceptron with 512-dimensional hidden layers, similar to the architecture from~\cite{deepsdf, metasdf, neuralpull, sal}. We emphasize, that our two-stage training pipeline is designed to be architecture-agnostic, and we provide results for alternate encoders and decoders in the supplement. For Eqs.~\eqref{metaeq} and \eqref{semieq} we the set the loss coefficients to $\lambda_{m} = \lambda_{s} = 0.1$, and $\lambda_{p}=0.01$ in Eq.~\eqref{selfeq}. We provide a full architecture description and additional training details in the supplemental document.

\paragraph{Experimental Setup} We source training and evaluation data from Acronym~\cite{acronym} and YCB~\cite{ycb}. Acronym is a post-processed subset of the popular ShapeNet~\cite{shapenet} dataset which contains 8,872 clean, synthetic 3D models in 262 shape categories. In contrast to ShapeNet, all of Acronym's meshes are \emph{watertight}, which ensures that they have well-defined SDFs. We use the 20 largest classes (2,995 meshes) as a labeled dataset $X$ for all fully-supervised methods, including our meta-learning first stage. The next 76 classes (3,408 meshes) form our unlabeled dataset $R$ for semi-supervised training. The remaining 166 classes (1,525 meshes) are withheld for testing. Preprocessing details can be found in the supplemental document. The YCB dataset~\cite{ycb} contains point clouds from 3D scans of common real-world objects and does not contain ground truth meshes. We use these point clouds as additional unlabeled test data to assess SDF predictions on out-of-distribution objects.

\paragraph{Evaluation} Given a point cloud $P$ that describes the surface of an object, the proposed method returns an implicit signed distance function $\Phi(P)$. We construct a cube with each dimension ranging from -1 to 1, then discretize it into a set of grids with resolution $256^3$. We run Marching Cubes~\cite{marching-cubes} at each grid point to reconstruct a mesh surface for qualitative assessment. For quantitative evaluation we measure the Chamfer Distance~\cite{pointnet} between 30,000 randomly sampled points on the surfaces of the reconstructed and ground truth meshes.

\paragraph{Experiments} We conduct experiments to investigate the generalizability and representational power of our method. Specifically, we analyze the effects of reconstructing seen and unseen objects (Sec.~\ref{sec:SeenUnseen}), self-supervision on unlabeled data (Sec.~\ref{sec:SelfVsUnsupervised}), performance on real-world point cloud scans (Sec.~\ref{sec:RealWorldPointClouds}), and ablation experiments (Sec.~\ref{sec:ablation}).

\begin{figure}[ht!]
    \includegraphics[width=\textwidth]{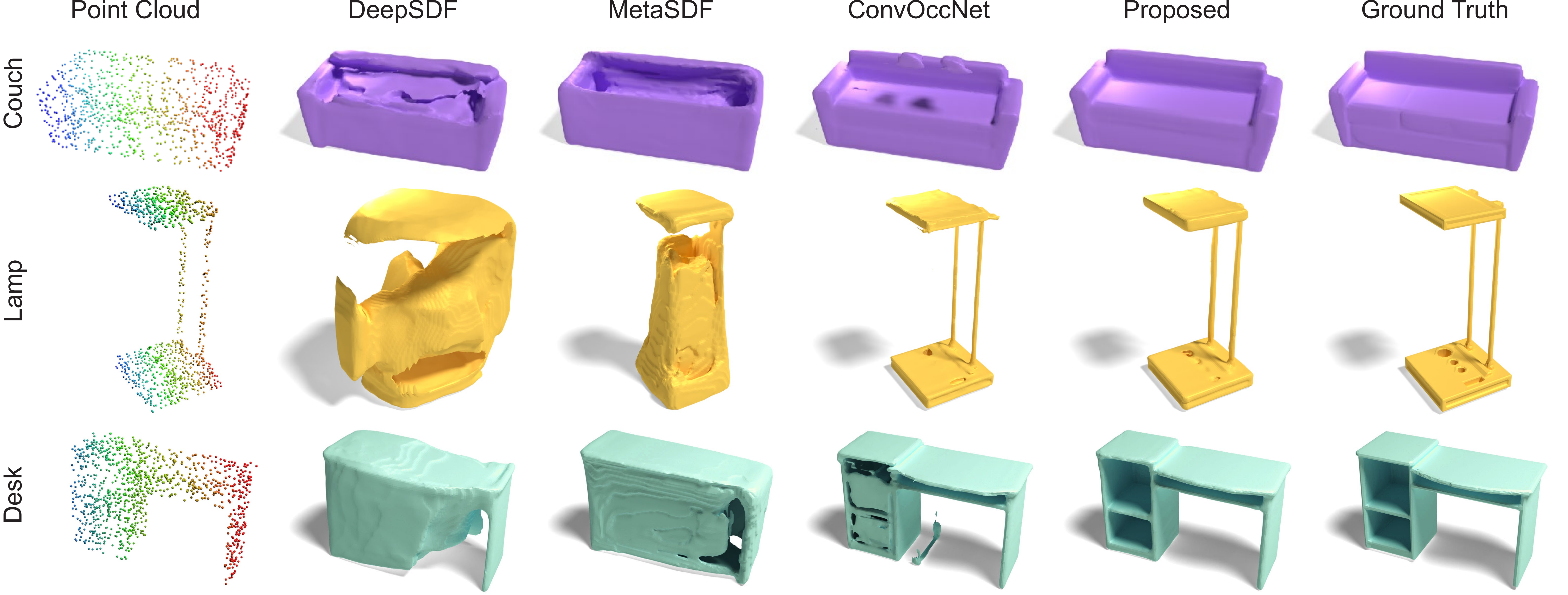}
    \caption{Reconstruction results for seen objects -- objects in the training set. DeepSDF and MetaSDF struggle to recover thin structures and non-convex surfaces due to their lack of learned shape priors. Compared to ConvOccNet our approach produces fewer artifacts in empty space.} 
    \label{fig:seen}
    \vspace{-2.5mm}
\end{figure}

\begin{figure}[ht!]
    \centering
    \includegraphics[width=\textwidth]{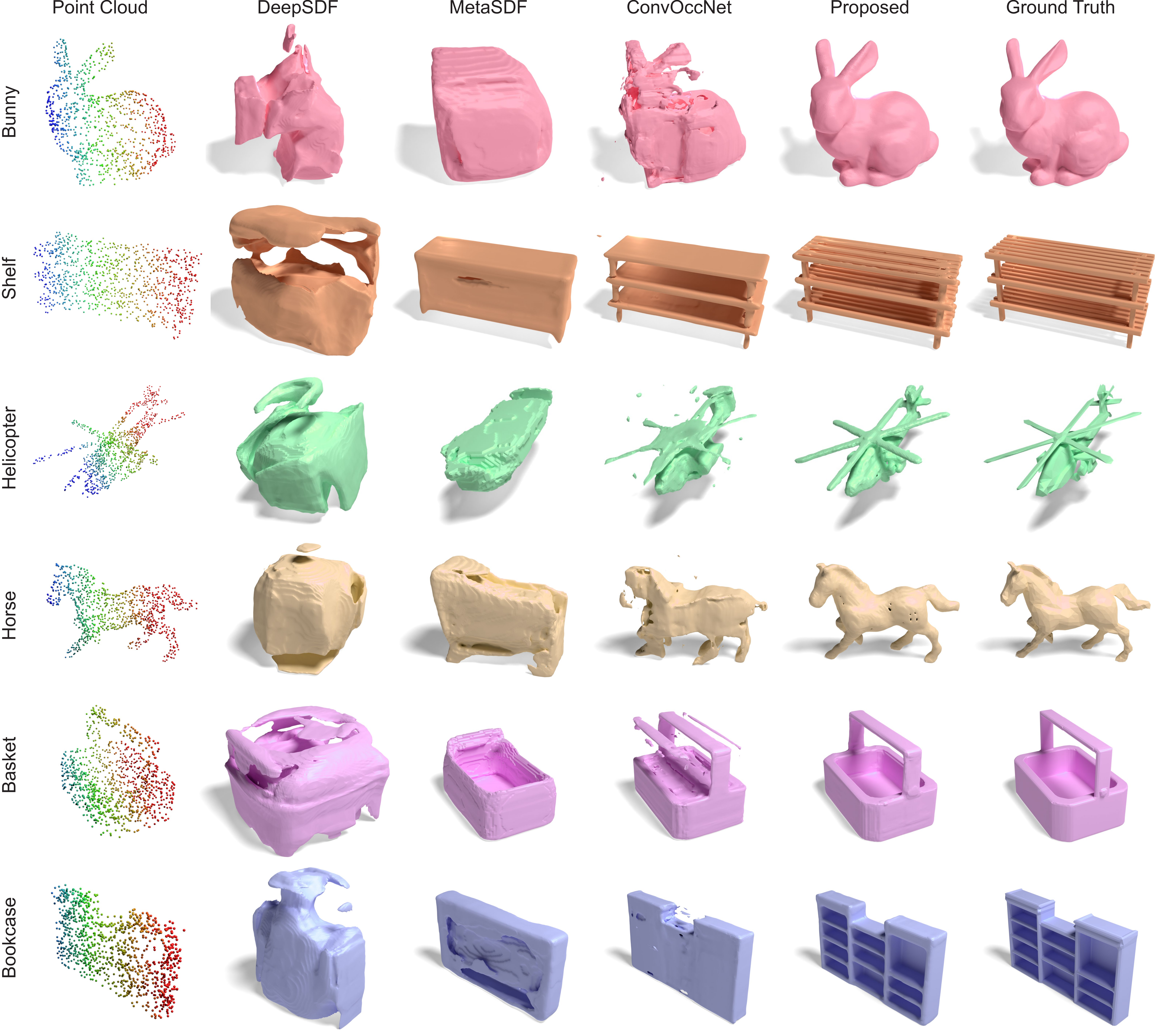}
    \caption{Reconstruction results for \emph{unseen categories} --  not included during training. Only our method recovers fine detail, such as the gaps in \emph{Shelf}'s surfaces and the curves of \emph{Bunny} and \emph{Horse}.} 
    \label{fig:unseen}
\end{figure}

\subsection{Reconstructing Seen and Unseen Shapes}
\label{sec:SeenUnseen}
In this subsection, we compare our method to existing approaches: DeepSDF~\cite{deepsdf}, MetaSDF~\cite{metasdf}, and Convolutional Occupancy Networks (ConvOccNet)~\cite{conv_occ} on both seen and unseen shapes.\footnote{Approaches such as~\cite{bacon, acorn, nglod} are designed to overfit to single objects, which significantly deviates from the focus of this paper, so we do not include comparisons with these methods.} 
For reconstructing seen data, we evaluate results on the 20 shared classes (see the experimental setup paragraph in Sec.~\ref{sec:experiments}). For testing on unseen classes, we use the remaining 166 categories that are \emph{disjoint} with any class used for training. To improve visual results, we apply refinement by fitting the model on the point cloud for a few iterations. Specifically, we employ our self-supervised loss without using any additional information beyond 5,000 raw input points. In contrast, DeepSDF~\cite{deepsdf} and MetaSDF~\cite{metasdf} use ground truth signed distance values for refinement. We include specific refinement details in our supplement.
Since ConvOccNet~\cite{conv_occ} does not perform refinement, we report quantitative results of our proposed method \emph{without} refinement for a fair comparison.

\paragraph{Seen Categories.}
Fig.~\ref{fig:seen} shows reconstruction results on seen shapes. We observe that existing methods and our approach are capable of reconstructing seen data. However, DeepSDF and MetaSDF fail to reconstruct thin structures such as the lamp poles due to the lack of a shape prior.
Quantitatively, reported in the middle column of Fig.~\ref{fig:hist}, even though we use the same encoder as ConvOccNet~\cite{conv_occ} to learn shape priors, our two-stage approach achieves a lower CD by an order of magnitude. 

\paragraph{Unseen Categories.}
When tested on unseen categories, as shown in Fig.~\ref{fig:unseen}, only our proposed method is able to reconstruct unseen categories, and accurately represents fine details such as narrow gaps in the Shelf example.
Quantitatively, as shown in the right column of the table in Fig.~\ref{fig:hist}, the proposed method achieves better performance. Notably, the gap between the CD of seen and unseen classes is lower in our case. While the CD of most methods double between seen and unseen classes, there is only a marginal increase for ours.

\subsection{Comparison to Unsupervised Methods on Unlabeled Data}
\label{sec:SelfVsUnsupervised}

\begin{wraptable}{r}{0.52\textwidth}
\vspace{-1em}
  \caption{Mean/median Chamfer Distance for self- and un-supervised methods when trained on 5 objects from the same \emph{QueenBed} class and 352 objects from 4 classes (\emph{QueenBed}, \emph{Plant}, \emph{TrashBin}, \emph{DeskLamp}). We report pre-refinement numbers for our approach, scale is $10^{-4}$.}
  \label{table:unlab}
  \centering
    \begin{tabular}{lcc}
        \toprule
        & Queen Beds & Four Classes \\
        \midrule
        SAL & 1.786 $/$ 1.869 &  65.55 $/$ 49.38\\
        NeuralPull & 1.745 $/$ 1.502 & 2.419 $/$ 1.812 \\
        Proposed (self) & 0.146 $/$ 0.139 & 0.218 $/$ 0.184 \\
        Proposed & \textbf{0.081} $/$ \textbf{0.077}  & \textbf{0.089} $/$ \textbf{0.061} \\
        \bottomrule
  \end{tabular}
  \vspace{-0.5em}
\end{wraptable}
A self/unsupervised method is essential for our semi-supervised stage to train on unlabeled data and for test-time refinement on only the raw input point cloud. In this subsection, we compare to existing unsupervised baselines: NeuralPull~\cite{neuralpull} and SAL~\cite{sal}. For a fair comparison, we use only our self-supervised loss in Eq.~\eqref{selfeq} to train directly on unlabeled data and do not perform test-time refinement (\emph{Proposed (self)} in Tab.~\ref{table:unlab}).  

Tab.~\ref{table:unlab} validates that when training a single Queen Bed class, our explicit sign penalty allows our self-supervised method to reconstruct sharper details from irregular surfaces. On the other hand, existing methods~\cite{sal, neuralpull} produce low losses for close approximations to the absolute values of the signed distances, rather than accurate sign predictions, resulting in loss of details where there is abrupt change in the surface (see supplement for visualizations). All non-supervised methods we test on, including ours, degrade when training degrade quickly as the number of meshes increases. 

The proposed method with semi-supervision scales well as we increase the number of classes. Thus, our method can tackle large-scale operations and take advantage of the availability of large amounts of unlabeled data. We also show in our supplement that using our self-supervised variant from this section leads to significantly more robust training.

\begin{figure}[t]
    \centering
    \includegraphics[width=\textwidth]{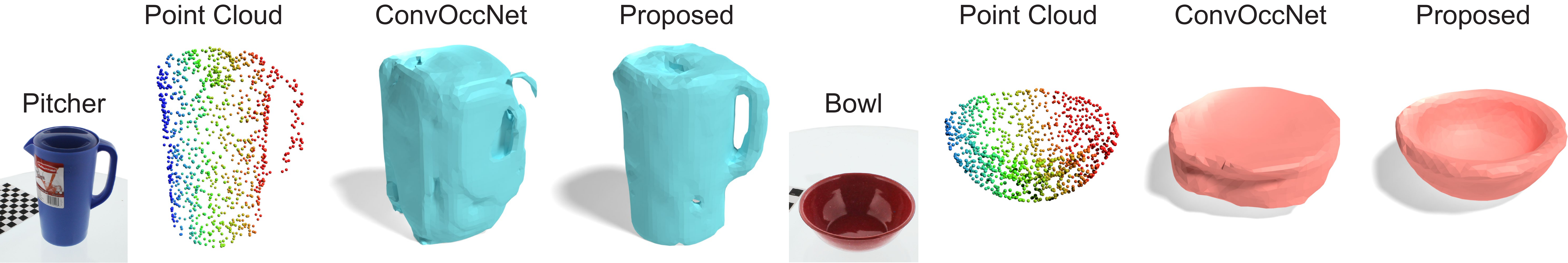}
    \caption{Reconstruction results for real scanned data from unseen categories in YCB~\cite{ycb}. This is a difficult setting, as the data has a different canonicalization, scaling, and distribution. Our proposed method is able to recover structures such as \emph{Pitcher}'s handle and the inside of \emph{Bowl}.} 
    \label{fig:ycb}
    \vspace{-1em}
\end{figure}
\begin{figure}[t]
    \centering
    \includegraphics[width=\textwidth]{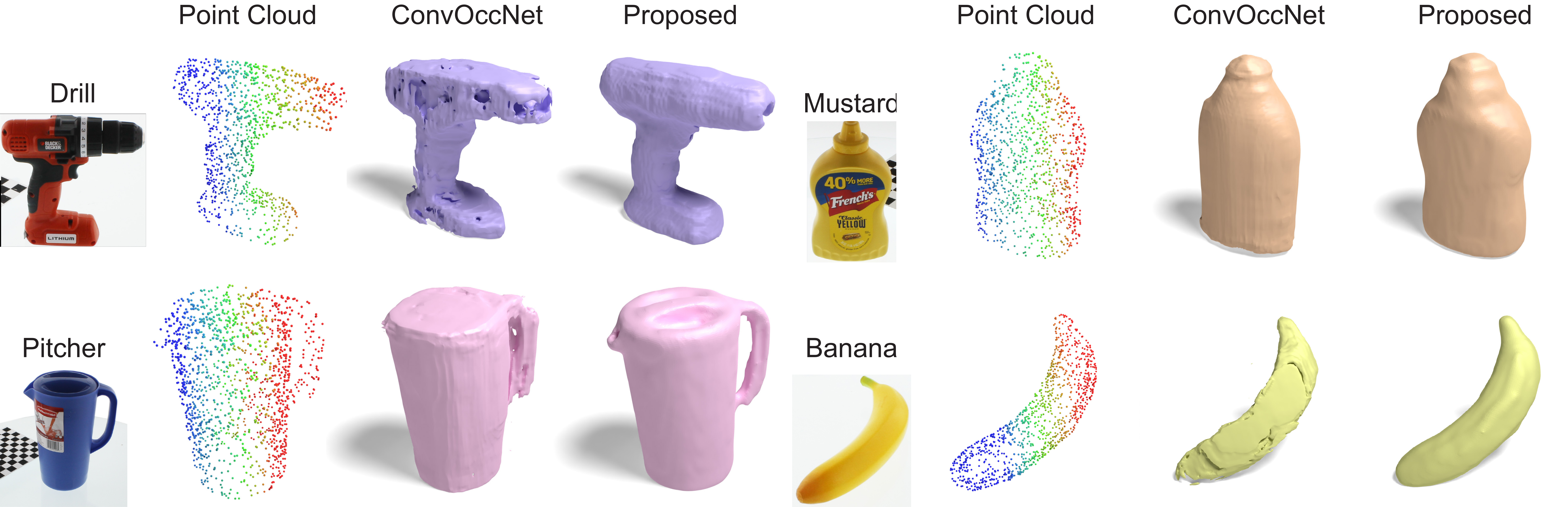}
    \caption{Reconstructions of YCB~\cite{ycb} after recentering and normalization of input point cloud.}
    \label{fig:more_ycb}
\end{figure}

\subsection{Reconstruction of Real-World Raw Point Clouds}
\label{sec:RealWorldPointClouds}
Next, we test our model on the YCB~\cite{ycb} dataset, which is a real-world point cloud dataset acquired from multi-view RGBD captures. The fused multi-view point clouds in this dataset resemble input measurements for a robotic part-picking or manipulation task. We demonstrate robust mesh reconstructions of the measured data, e.g., recovering the "handle" of a pitcher in Fig.~\ref{fig:ycb} and \ref{fig:ycb}, which may serve as input to complex robotic grasping tasks. 

YCB is a different domain from Acronym~\cite{acronym} and we show two experiments to thoroughly test performance of out-of-distribution data. First, we perform no scaling or normalization of the input point cloud, shown in Fig.~\ref{fig:ycb}. Then, we also recenter and normalize the input point cloud (which we would do in a real-world setting) and show results in Fig.~\ref{fig:more_ycb}. We omit results from DeepSDF~\cite{deepsdf} and MetaSDF~\cite{metasdf} since they require ground truth signed distance values during test time. As shown in Fig.~\ref{fig:ycb} and \ref{fig:more_ycb}, ConvOccNet~\cite{conv_occ} loses detail such as the handle of the pitcher. Our proposed method is able to generate 3D shapes with detail and plausible geometry, even generalizing to different scaling, illustrating its ability of handling in-the-wild and out-of-distribution data. 

To further understand what level of noise our method is susceptible to, we gradually add Gaussian noise with mean zero and variance $\sigma ^2$ to the \emph{Pitcher} point cloud in YCB~\cite{ycb} and evaluate its CD. Both quantitatively and qualitatively, our method is able to adapt to noise with $\sigma ^2 < 0.5$, but afterward the reconstruct object degrades severely. We show results in Table ~\ref{table:noise}. In the future, adding more diverse unlabeled data to the training of the proposed model, including different domains, scaling, and noise, may further boost the generalizability to any target data. 
\begin{table}[h]
  \caption{Adding Gaussian noise with zero mean and variance $\sigma ^2$ to the \emph{Pitcher} point cloud in YCB and evaluating Chamfer Distance (CD), scale is $10^{-4}$. }
    \label{table:noise}
    \centering
    \begin{tabular}{ccccccc}
    \toprule
       $\sigma ^2$ & 0.0&0.01&0.05&0.1&0.15&0.2\\
    \midrule
    CD  & 0.870&0.911&3.122&5.208&7.429&7.391   \\
    \bottomrule
    \end{tabular}
\end{table}

\subsection{Ablation Experiments}
\label{sec:ablation}

In this subsection, we analyze individual components of our two-stage learning. We show quantitative results in Tab.~\ref{table:ablation}. \emph{Only meta} refers here to the meta-learning first stage on 20 classes. This outperforms existing fully supervised methods (see Fig.~\ref{fig:hist}) for both seen and unseen data, while using the same amount of data and no additional manual annotations. \begin{wraptable}{r}{0.49\textwidth}
  \vspace{1em}
  \caption{Mean/median Chamfer Distance (CD) for various ablations on 20 classes of seen data and 166 classes of unseen data. We report pre-refinement numbers, scale is $10^{-4}$.}
  \label{table:ablation}
  \begin{tabular}{lcc}
    \toprule
          & Seen Classes & Unseen Classes\\
    \midrule
    Proposed & \textbf{0.351}$/$\textbf{0.166}  & \textbf{0.407}$/$\textbf{0.250}    \\
    \midrule
    Only meta  & 0.503$/$0.218  & 0.515$/$0.297     \\
    Only semi & 0.821$/$0.510  & 0.889$/$0.612     \\
    \midrule
    Sup. 20 cls &  0.695$/$0.468 & 1.164$/$1.367     \\
    Sup. 76 cls &  0.713$/$0.550 & 0.908$/$0.866     \\
    
    \bottomrule
  \end{tabular}
  \vspace{-0.5em}
\end{wraptable}
The favorable results validate the generalization capabilities of our episodic training scheme. Our semi-supervision stage further improves reconstruction quality by introducing more shapes without requiring annotations. \emph{Only semi} refers to only training the proposed semi-supervision stage from scratch on all classes without transferring the learned representation from the meta-learning stage. Even though this stage has access to both labeled and unlabeled data, it has a higher CD compared to our final two-stage approach. This is because errors in the proxies of unlabeled data accumulate and result in training that is less robust. This indicates that our meta-learning stage indeed builds a strong foundational prior that ensures the second stage semi-supervised training is robust.

\emph{Sup. (20 cls)} refers to training on the labeled set of 20 classes without episodic training and resplitting. \emph{Sup. (76 cls)} refers to training on 76 classes with ground truth annotations but without episodic training and resplitting. The CD for seen classes is similar to the \emph{Only meta} setting, but our meta-learning method outperforms both baselines by a large margin when it comes to reconstructing unseen classes, despite being trained with fewer categories.

\section{Conclusion}
\label{sec:conclusion}
In this work, we devise a two-stage semi-supervised meta-learning approach that achieves strong generalization to unseen object categories. This method is able to fit to any target point cloud without additional annotations and outperforms existing SDF methods in both seen and unseen categories. Our method is not without limitations, which we plan to address in the future. Similar to existing SDF representations~\cite{conv_occ}, we do not achieve real-time inference due to large number of model parameters. Learning from large-scale multi-object point clouds, with abundant datasets for automotive scenes \cite{kitti_auto, waymo} is also an exciting future direction. The proposed approach may also serve as a representation for inverse rendering methods that tackle vision tasks in an analysis-by-synthesis approach.

\section*{Acknowledgments}
We would like to thank Hong-Yuan Liao and Chien-Yao Wang from Academia Sinica, Taiwan for constructive discussions and advice in completing this work. Ilya Chugunov was supported by an NSF Graduate Research Fellowship. Felix Heide was supported by an NSF CAREER Award (2047359), a Sony Young Faculty Award, a Project X Innovation Award, and an Amazon Science Research Award.

\bibliographystyle{unsrt}
\bibliography{bib.bib}



\section*{Supplementary material}

# GenSDF
# Supplementary Information

**Gene Chou**
Princeton University
gchou@princeton.edu

**Ilya Chugunov**
Princeton University
chugunov@princeton.edu

**Felix Heide**
Princeton University
fheide@princeton.edu

## Contents

## 1   Implementation Details

In this section, we provide further details for reproducibility, including data preprocessing, training steps and hyperparameters, and model architecture. We will also release all code to facilitate experiments.

### 1.1   Data Processing

**Preprocessing**   We recenter and normalize all objects in the Acronym [1] dataset. We use the normalized object coordinate system (NOCS) [2]. This means we contain all points in 3D space in a cube and uniformly scale each object such that the diagonal of its tight bounding box has a length of 1 and is centered within the cube. Different from the original definition [2] where the cube

has coordinates ranging from (0,0,0) to (1,1,1), we set our cube to have coordinates ranging from (-1,-1,-1) to (1,1,1). That is, the length of each side on our cube is 2 instead of 1.

We do not perform this preprocessing for the YCB [3] dataset which we use for testing. The point clouds are obtained from real-world scans and have size-accurate scaling. This makes testing difficult since the scaling is out-of-distribution.

**Labeled Data** For labeled data, we largely follow [4] for preprocessing. From each mesh we sample 235,000 points on the surface. These points form a point cloud $P = \{p_i \in \mathbb{R}^3\}_{i=1}^{235000}$ that describe the surface and have signed distances of 0. Then, for each point $p_i$, we sample two Gaussian distributions with mean 0 and standard deviation 0.005 and 0.0005, respectively. We add these distributions to $p_i$ to obtain two query points per surface point. Since we have the mesh, we can calculate ground truth signed distance values for supervised training and we store all points and corresponding signed distances, one per object. During one training step of one mesh, we randomly sample 1,024 surface points for generating shape features through the encoder, and sample 16,000 query points for predicting signed distance values. We did not run a grid search for determining the optimal number of points. We experimented with a larger number for generating shape features (e.g., 5,000 instead of 1,024), which led to a decrease in generalization capabilities.

**Unlabeled Data** For unlabeled data, we largely follow the preprocessing from [5] for generating points on and near the surface. For each input point cloud, we randomly sample 5,000 points and obtain $P = \{p_i \in \mathbb{R}^3\}_{i=1}^{5000}$. Then for each point $p_i$, we sample 20 Gaussian distributions each with mean 0 and standard deviation $\sigma$, where $\sigma$ is the distance between $p_i$ and its 50-th nearest neighbor. We add each distribution to $p_i$ to obtain 20 query points near the surface. Next, to ensure the model does not overfit to generated queries that only take up a small spatial proportion of the cube, we sample 30,000 additional points within the cube with each dimension ranging from -1 to 1. We then store the nearest neighbor $p$ to each query point $x$ by finding the shortest distance between $x$ and $p \in P$. In total, we have 5,000 points on the surface and 130,000 query points near the surface. During our first meta-learning stage, we randomly sample 1,024 surface points and 16,000 query points each training step, same as labeled data. During the second stage of semi-supervised training, we randomly sample 5,000 surface points and 5,000 query points each training step. More importantly, in this stage we train all 130,000 query points for each unlabeled object before moving on to the next one; that is, we train each unlabeled object in 130,000 / 5,000 = 26 steps. In [5], 20,000 points per input point cloud are sampled and 25 query points per surface point, but we reduce this number by a fraction to speed up training due to the higher number of training meshes.

## 1.2 Training Details

**Model Hyperparameters** For Eqs. (3) and (9) (main paper) we the set the loss coefficients to $\lambda_m = \lambda_s = 0.1$, and $\lambda_p = 0.01$ in Eq. (7). $\lambda_m$ and $\lambda_s$ determine the weights of the unlabeled data and we found that $0.1$ was a good tradeoff between learning from unlabeled data and still mainly guiding the model with supervised, labeled data. A higher value such as $1.0$ would lead to accumulation of inaccuracies in the proxies of unlabeled data. The parameter $\lambda_p$ determines the weight of points on the surface vs points near the surface for unlabeled data. Here, setting $\lambda_p = 0.01$ helped with thinner structures, but a larger value led to gaps and holes in the reconstructions.

For our meta-learning stage, we define one epoch as sampling one object from one class. Since our labeled dataset consists of 20 classes, 20 objects are seen every epoch. We re-split the dataset into 10 labeled sets and 10 unlabeled sets every 1000 epochs and train for 40,000 epochs. We did not run search to determine the optimal value of split frequency or split ratio.

**Training Hyperparameters** We use PyTorch [7] and train all models with the ADAM [8] optimizer. We set learning rates to $1 \times 10^{-4}$ with no decay.

For our meta-learning stage, we use 2995 meshes from 20 classes and the max batch size is 10 (each step trains one labeled and one unlabeled sample), so we set the batch size to 10. This takes up roughly 4.5 GB of GPU memory and we trained to 40,000 epochs in 2 days on an NVIDIA A100 GPU.

For our semi-supervised stage, one epoch is defined as sampling all objects once; i.e., each epoch iterates through 2995 labeled and 3408 unlabeled objects (see main paper for our specific data split).

Table 1: Decoder architecture. Following DeepSDF [4] with a number of changes, see text.

| Layer | In-Features | Out-Features | Notes |
|---|---|---|---|
| Linear | 259 | 512 | input = shape code (256) + xyz (3) |
| ReLU | | | |
| Linear | 512 | 512 | |
| ReLU | | | |
| Linear | 512 | 512 | |
| ReLU | | | |
| Linear | 512 | 512 | |
| ReLU | | | |
| Linear | 771 | 512 | skip connection = 512 + 259 |
| ReLU | | | |
| Linear | 512 | 512 | |
| ReLU | | | |
| Linear | 512 | 512 | |
| ReLU | | | |
| Linear | 512 | 512 | |
| ReLU | | | |
| Linear | 512 | 1 | output is not regressed with any activation |

Table 2: Encoder architecture. Following ConvOccNet [6]. "+" represents a stacking of operations. "lin" represents a fully-connected layer. "conv(a,b,c)" represents a 2D convolutional layer with kernel size a, stride b, padding c, and convT is a transposed convolution. "pool(a,b,c)" represents a 2D max pool with kernel size a, stride b, padding c.

| Layer Name | Type | In-/Out-Features |
|---|---|---|
| Linear | lin | 3/128 |
| ResnetBlock1 | lin+ReLU+lin+ReLU+shortcut | 128/128 |
| ResnetBlock2 | lin+ReLU+lin+ReLU+shortcut | 128/128 |
| ResnetBlock3 | lin+ReLU+lin+ReLU+shortcut | 128/128 |
| ResnetBlock4 | lin+ReLU+lin+ReLU+shortcut | 128/128 |
| ResnetBlock5 | lin+ReLU+lin+ReLU+shortcut | 128/64 |
| Linear | lin+ReLU | 64/256 |
| DownConv1 | conv(3,1,1)+conv(3,1,1)+pool(2,2,0) | 256/32 |
| DownConv2 | conv(3,1,1)+conv(3,1,1)+pool(2,2,0) | 32/64 |
| DownConv3 | conv(3,1,1)+conv(3,1,1)+pool(2,2,0) | 64/128 |
| DownConv4 | conv(3,1,1)+conv(3,1,1) | 128/256 |
| UpConv1 | convT(2,2,0)+conv(3,1,1)+conv(3,1,1) | 256/128 |
| UpConv2 | convT(2,2,0)+conv(3,1,1)+conv(3,1,1) | 128/64 |
| UpConv3 | convT(2,2,0)+conv(3,1,1)+conv(3,1,1) | 64/32 |
| Final Conv | conv(1,1,0) | 32/256 |

Note that as mentioned previously, we train on all 130,000 query points for each unlabeled object. We train for 500 epochs. We set the batch size to 128 and train on an NVIDIA A100 for 5 days.

**Model Architecture** As described in the main paper, we use the encoder from [6] and the decoder from [4]. For the encoder, we set the latent size to 256 and hidden dimensions to 64. We use the modified PointNet [9] proposed in [6] that uses plane projection to learn local geometric features, and parallel UNets [10] to aggregate spatial information. For the decoder, we set the number of layers to 8, and use 512 hidden dimensions. We use ReLU activation after each layer except the last, and add a skip connection to the fourth layer. Different from [4], we do not use any normalization and we do not regress the final output with a Tanh activation. We also apply the geometric initialization

3

proposed in [11]. We provide a detailed list in Tab. 1 and 2. We also evaluate different architectures in Sec. 3.

### 1.3 Test-time Refinement

Most existing works that fit to multiple objects perform test-time refinement [4, 12, 11]. DeepSDF [4] creates a new latent code in 800 iterations while freezing the decoder. MetaSDF [12] performs 5 gradient steps. Notably, both methods use *ground truth signed distance* values for refinement. MetaSDF [12] showed some experiments of fitting only on point clouds, but reconstructions lack detail especially for intricate shapes.

The proposed self-supervised method allows us to perform test-time refinement on only the point cloud and no additional input. We sample 5,000 points per input point cloud. Sampling more points here, such as 15,000, leads to more detailed reconstruction outputs though the difference is not significant. Sampling size can be adjusted based on availability of data.

Using the 5,000 points, we then sample 20 queries per point as well as an additional 30,000 queries from the cube using the same method as in Sec. 1.1. Again, we emphasize that all these points are generated from the input point cloud and the process does not involve additional ground truths or annotations. Similar to DeepSDF [4], we perform refinement for 800 iterations. Each iteration, we randomly sample 5,000 query points. We set the learning rate to $1 \times 10^{-4}$. Finally, as we discuss in more detail in Sec. 5, we set the reconstruction level set to 0.005 due to inaccuracies in the unlabeled proxies (we set to 0.0 without refinement).

Our model is *also capable of reconstructing any target point cloud directly, that is without refinement,* and in the main paper we report numbers without refinement. We sample 1,024 points per point cloud, the same number we use during training. We show visualizations of reconstructions without refinement in Fig. 4. Our method still reconstructs details well although refinement is optimal for intricate shapes.

## 2 Additional Results

In the main paper we report numerical results of testing on 166 unseen categories. In supplemental Fig. 1, 2, and 3 we provide additional visualization results. Our testing method for both seen and unseen categories is the same as in Fig. 4 and 5 from the main paper but here we show a more diverse set of classes and shapes, including thin and hollow structures.

## 3 Alternative Network Architectures

We experiment with other network architectures and show visualizations in Fig. 4. First, we use the SIREN [13] architecture for our decoder. Since our point clouds only contain 3D coordinates and no normals, we exclude the loss term that requires normals. We also experiment with encoding coordinates with Fourier Features [14]. These experiments validate that our learning method is indeed applicable to different architectures, and, at the same time, validates the proposed architecture choice in our specific method.

## 4 Single-Object Methods

We next discuss single object methods which have been recently popular. We note that single object methods [15, 16, 17] solve a different problem than the proposed method. These methods focus on reconstructing single objects and scenes with detail, rather than generalization to multiple objects.

Nevertheless, we include an example of Ma et al. [17] and show results in Fig. 5. This method (Predictive Context Priors) reconstructs a *single unlabeled point cloud* with fine detail, including the gaps between the launchers at the side of the helicopter, but comes at the expense of learning shape priors and generalization capabilities. For completeness, we show a reconstruction of training three point clouds at once, though we note that this is not an intended setting for single object methods such as this method.

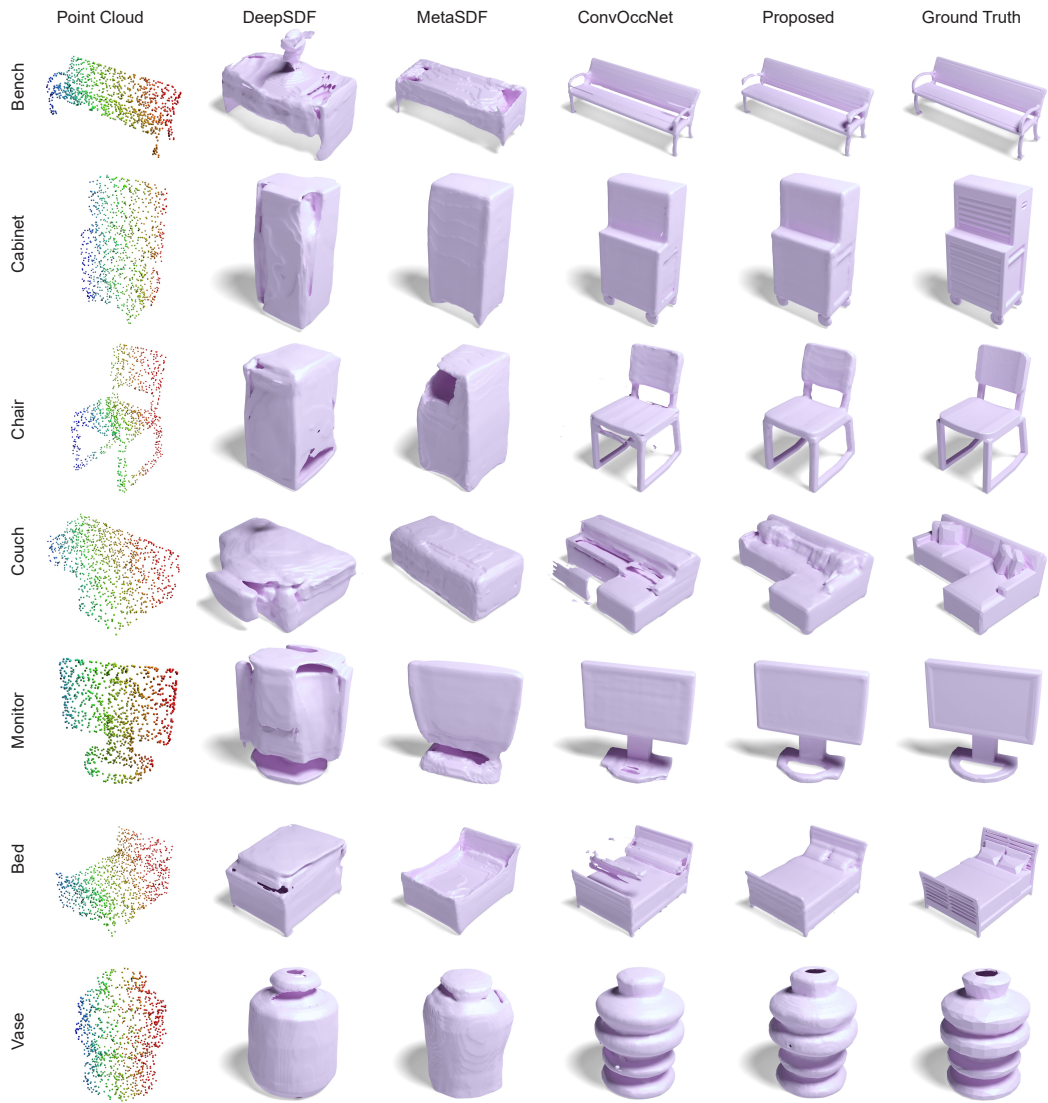

Figure 1: Additional reconstruction results on *seen* objects.

5

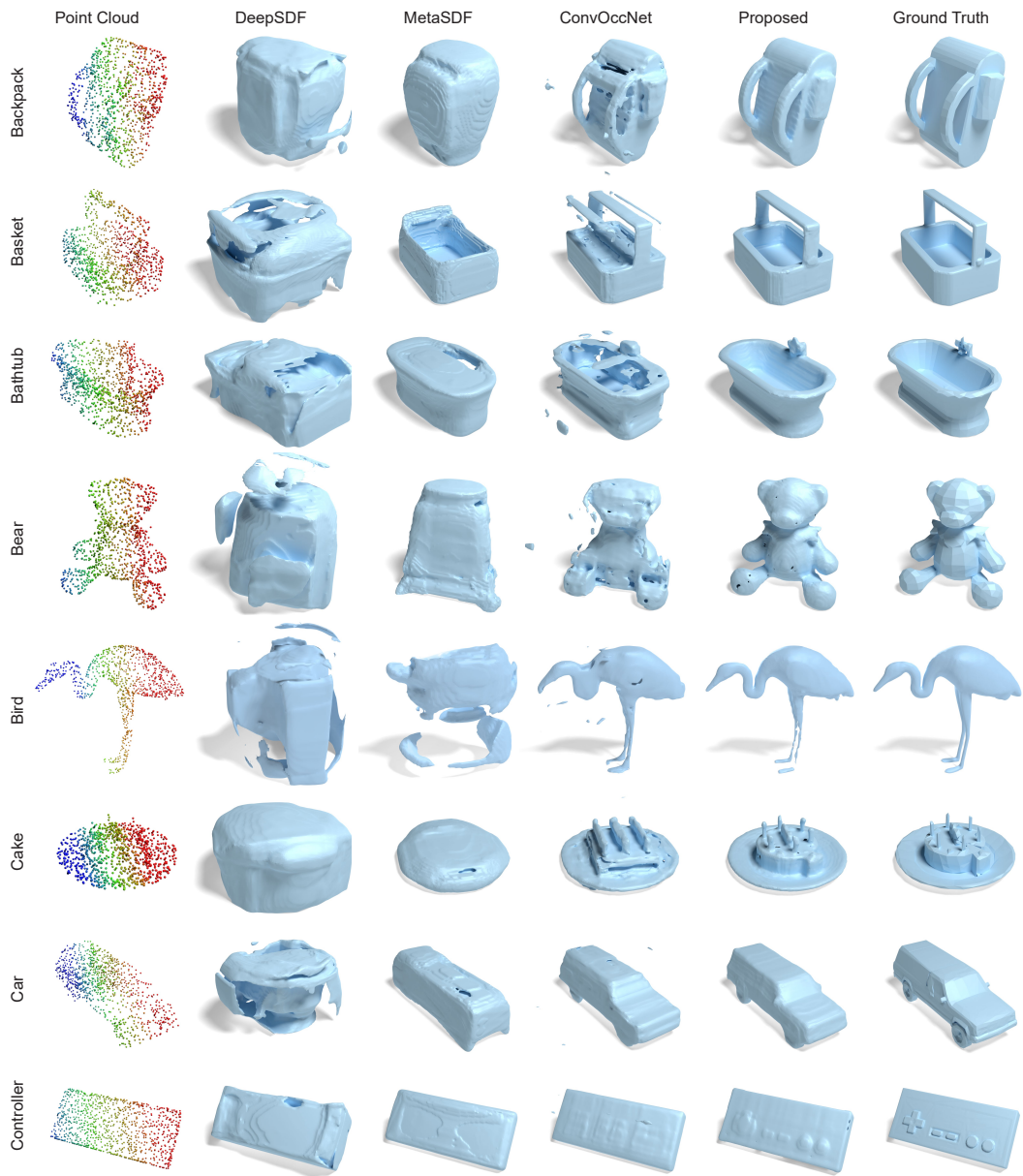

Figure 2: Additional reconstruction results on objects from *unseen* categories.

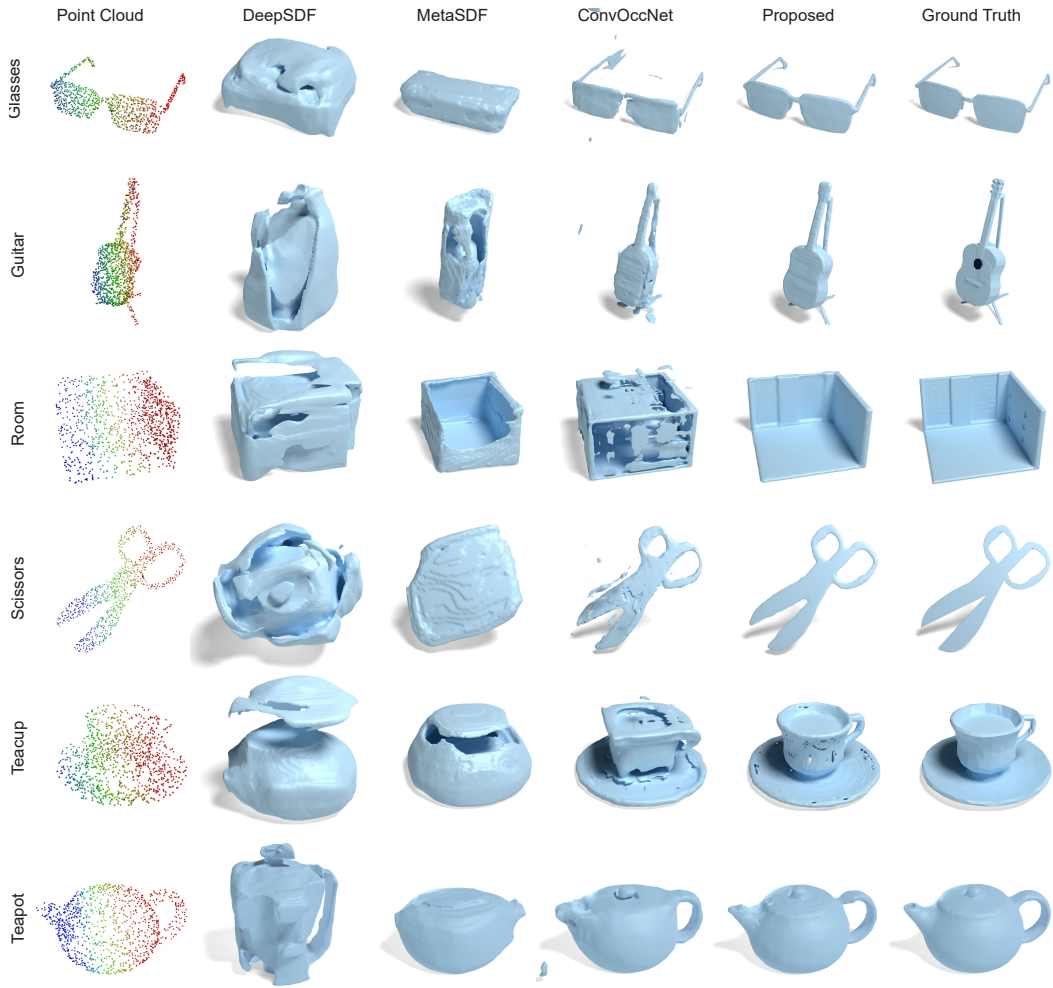

Figure 3: Even more reconstruction results on objects from *unseen* categories.

However, even for a single point cloud, Predictive Context Priors takes 30,000 input points and more than 10,000 iterations to train. On the other hand, our method reconstructs with high quality after just 800 iterations and 5,000 input points.

The main focus of Predictive Context Priors [17] and many other recent papers [15, 16] is different but complementary to ours. We leave the exploration of jointly achieving fine details with generalization across shapes as potentially exciting future work.

## 5 Unsupervised Methods for Training Unlabeled Data

Next, we analyze reconstructions of unlabeled data using our self-supervised method in comparison to existing unsupervised baselines: NeuralPull [5] and SAL [11]. The top row of Fig. 6 shows results of training and reconstructing objects from a single QueenBed class with 5 meshes. As a result of the explicit sign penalty, our method is able to reconstruct sharper details from irregular surfaces such as pillows. On the other hand, existing methods produce low losses for close approximations to the absolute values of the signed distances, rather than accurate sign predictions. Therefore, they tend to reconstruct a smoother curvature on shapes.

Using this experiment, we dive into a deeper comparison between our self-supervised loss and the approach described in NeuralPull [5]. First, we report accuracies of sign predictions between NeuralPull and our self-supervised method. We run inference on 130,000 points near the surface of

7

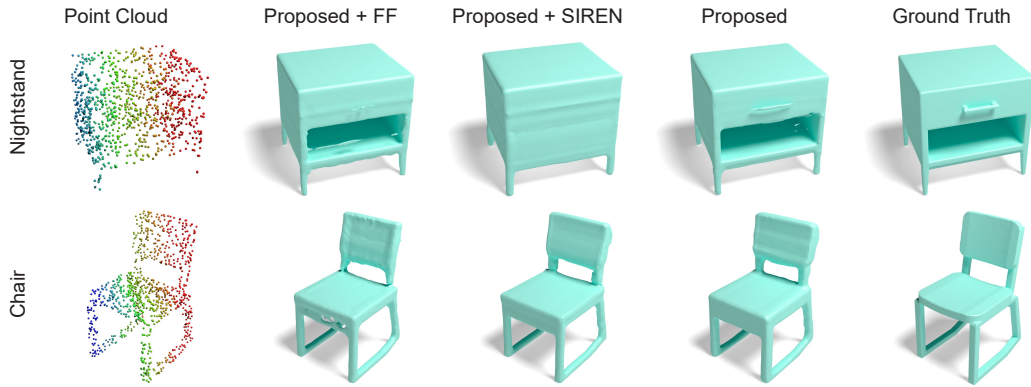

Figure 4: Testing on alternative architectures, including encoding coordinates using Fourier Features [14] and using the decoder of SIREN [13]. Results are *without* optimization or test-time refinement. These experiments validate that our learning method is applicable to different architectures.

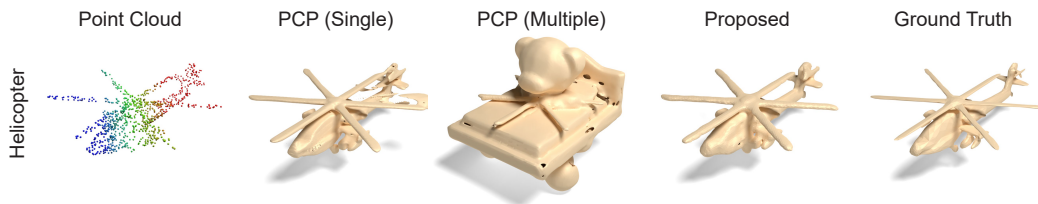

Figure 5: Comparing results to Predictive Context Priors (PCP) from Ma et al. [17]. PCP is designed to fit to a single object or scene with detail, taking 30,000 input points and 10,000 iterations. Our proposed method performs refinement using 5,000 input points and 800 iterations. PCP is not intended for training on multiple objects; we include PCP (multiple) which is trained on 3 objects only for completeness.

each of the five QueenBed point clouds (130,000 * 5 = 650,000 total). While both methods produced low $\ell_2$ losses, in Tab. 3 our method achieves higher accuracy in sign prediction.

Table 3: Confusion matrices for NeuralPull and our self-supervised loss sign predictions on five point clouds of the QueenBed class. Of the five point clouds, there are a total of 650,000 points to predict, with 449,650 being positive and 200,350 being negative.

| NeuralPull | Pred Positive | Pred Negative |
|---|---|---|
| 449,650 Pos | 37,779 (0.974) | 11,871 (0.026) |
| 200,350 Neg | 37,986 (0.190) | 162,363 (0.810) |

| Our SSL | Pred Positive | Pred Negative |
|---|---|---|
| 449,650 Pos | 449,245 (0.999) | 405 (0.001) |
| 200,350 Neg | 2,765 (0.014) | 197,585 (0.986) |

Despite good reconstruction results with small amounts of data, no purely self-supervised method works well when training on multiple classes. We randomly select 4 classes and 352 meshes and find that both NeuralPull [5] and the purely self-supervised variant of our method degrade quickly as the number of meshes increases. Interestingly, in this setting we found that the errors of using nearest neighbors rather than ground truth signed distance values accumulate and distort the original surface boundary. For fair comparison, we follow NeuralPull and use 0.005 level-set (rather than zero level-set) to represent the surface of the object. We assume the deviation in the positive direction a result of more training samples that are outside the surface since most surfaces we train on are convex.

8

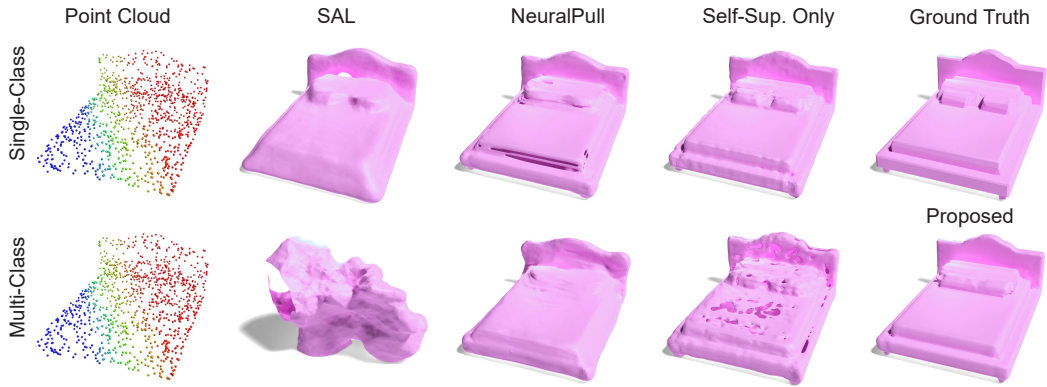

Figure 6: Comparing self/unsupervised methods. Our self-supervised loss penalizes incorrect signs and leads to sharper details from irregular surfaces such as pillows. Neither method works well when multiple categories are introduced. Our full proposed method includes labeled data for supervision which allows for scalability to multiple classes. All results are without test-time refinement.

This issue occurs also in Predictive Context Priors [17] which is unsupervised. Slight deviation from the zero level-set seems to be a problem with self/unsupervised methods; we leave this for future work. When trained with labeled data in our two-stage approach, we were able to reconstruct based on the zero level-set. However, after test-time refinement, the 0.005 level-set was required.

This supports our finding that methods without supervision are insufficient for tackling large-scale operations and are unable to take advantage of the availability of large amounts of unlabeled data. However, empirically we found that using our self-supervised method compared to existing unsupervised methods was more robust when trained along labeled data in our two-stage approach. When using NeuralPull [5] for unlabeled data, the model could not converge.

## 6  Effectiveness of Meta-Learning a Prior

In the main text, we mention there are a few advantages of meta-learning a prior. Compared to directly training on labeled and unlabeled data, our meta-learning process provides additional flexibility as we can tune the split frequency and ratio of labeled and "pseudo-unlabeled" data. The model repeatedly sees the same point clouds with and without labels, so shape features generated during labeled splits are shared with "pseudo-unlabeled" splits. In contrast, we find that semi-supervised training from scratch is unstable because errors in unlabeled data accumulate. Here, we provide two experiments to illustrate the effectiveness and limitations of our meta-learning approach.

**Training on diverse vs similar classes**  We train two models on our meta-learning first stage. The first trains on the following six diverse shape classes, with a mix of small, large, planar, convex, and concave surfaces: 'Bench', 'QueenBed', 'EndTable', 'FloorLamp', 'Monitor', 'PottedPlant.' The second trains on six classes that are semantically and geometrically similar, large planar shapes: 'EndTable', 'AccentTable', 'CoffeeTable', 'DiningTable', 'RoundTable', 'Table.' Then we evaluate on 166 unseen classes. The first model outperforms the second in reconstruction quality, validating that the diversity in the training set is more important than the raw number of classes. We do note that the difference in CD is noticeable but not very significant. This validates the effectiveness of our meta-learning approach: even with less diverse data, our model can still produce generalized priors. We provide quantitative results in Tab. 4.

Table 4: Results of training on six diverse and six similar classes. Mean/median Chamfer Distance (CD) of 166 unseen classes, scale is $10^{-4}$.

|  | Six Diverse Classes | Six Similar Classes |
| --- | --- | --- |
| CD | 0.972 / 0.657 | 1.321 / 0.934 |

**Training only with the supervised loss**   We train two separate models for a prior. Both train on six classes: 'Bench', 'QueenBed', 'EndTable', 'FloorLamp', 'Monitor', 'PottedPlant,' with a total of 600 meshes. The first model is trained using our two-stage learning approach. The second model is trained with only the supervised loss, but introduces dropout (20%) and small amounts of Gaussian noise perturbations to the point clouds to reduce overfitting.

Then, we use both priors to train our second stage in exactly the same fashion. With our prior, our model performs significantly better when reconstructing unseen classes. We provide quantitative results on 166 unseen classes in Tab. 5. We highlight that after training stage 1, both methods perform on par on unseen classes. However, our prior that emulates semi-supervised training is able to significantly outperform in the second stage, where we introduce true unlabeled data samples.

Table 5: Comparison of our approach to a prior trained only with the supervised loss. Mean/median Chamfer Distance (CD) of 166 unseen classes, scale is $10^{-4}$.

|  | Proposed | Supervised Prior (Augmentation + Dropout) |
|---|---|---|
| After stage 1 | 0.972 / 0.657 | 1.027 / 0.697 |
| After stage 2 | 0.420 / 0.316 | 0.851 / 0.568 |

# 7  Reproductions and Licenses

All datasets we use and existing methods we compare to have released their code as open source. For fair comparison, we make no changes to released code except for the dataloader for loading our datasets. We test on the same hardware platform. We will also release our own code as open source under the MIT license.

# 8  Limitations

There are a few limitations of our work that we would like to address in future work. First, we need multiple categories to learn a diverse set of geometry. Training on just one category (e.g., chairs) will not allow the model to generalize to multiple intricate shapes. Each category should also contain a certain number of representative objects. We did not run experiments to determine the number of categories required, but in our work we choose 20 categories, each with at least 80 objects, which worked out well.

Second, for intricate details, test-time refinement is required. The downside of test-time refinement is that each object requires more time to reconstruct. Fortunately, our self-supervised method allows for refinement without additional inputs other than the point cloud.

Finally, our method trains and tests on full-view point clouds. In future work, we would experiment with single or partial-view point clouds for shape completion tasks.

# 9  Broader Impact

Our method has no direct ethical impact, but potentially could be used for 3D reconstruction tasks in face recognition and human reconstruction, which could result in privacy issues or gender and racial biases.



\end{document}